# Dual-Mode Deep Anomaly Detection for Medical Manufacturing: Structural Similarity and Feature Distance

Julio Zanon Diaz, George Siogkas, and Peter Corcoran, *Fellow, IEEE.*

***Abstract*— Automated visual inspection in medical-device manufacturing faces unique challenges, including extremely low defect rates, limited annotated data, hardware restrictions on production lines, and the need for validated, explainable artificial-intelligence systems. This paper presents two attention-guided autoencoder architectures that address these constraints through complementary anomaly-detection strategies. The first employs a multi-scale structural-similarity (4-MS-SSIM) index for inline inspection, enabling interpretable, real-time defect detection on constrained hardware. The second applies a Mahalanobis-distance analysis of randomly reduced latent features for efficient feature-space monitoring and lifecycle verification. Both approaches share a lightweight backbone optimised for high-resolution imagery for typical manufacturing conditions. Evaluations on the Surface Seal Image (SSI) dataset—representing sterile-barrier packaging inspection—demonstrate that the proposed methods outperform reference baselines, including MOCCA, CPCAE, and RAG-PaDiM, under realistic industrial constraints. Cross-domain validation on the MVTec-Zipper benchmark confirms comparable accuracy to state-of-the-art anomaly-detection methods. The dual-mode framework integrates inline anomaly detection and supervisory monitoring, advancing explainable AI architectures toward greater reliability, observability, and lifecycle monitoring in safety-critical manufacturing environments. To facilitate reproducibility, the source code developed for the experiments has been released in the project repository, while the datasets were obtained from publicly available sources.**

***Note to Practitioners*— This work was motivated by the need to replace human visual inspection in medical-device manufacturing with reliable, traceable automation. The proposed approach combines two complementary anomaly-detection modes within one deep-learning framework. The structural-similarity mode performs inline defect detection in real time, while the feature-distance mode enables supervisory monitoring of process stability and model drift. Both modes operate effectively with limited data and can run on single-GPU industrial workstations.**

**The dual-mode architecture integrates naturally with existing automation infrastructure in two layers. At the edge, the attention-guided backbone and structural-similarity detector run on validated industrial hardware, providing rapid part classification and meeting inline quality-control requirements. At the distributed or cloud level, features are transmitted through industrial protocols such as MQTT, batched with MES metadata (timestamps, unit identifiers, and batch information). Using these feature distributions and Mahalanobis distances, the system performs batch-to-batch drift analysis and other advanced analytics. When drift is detected, reconstructed images can be reviewed to determine whether quality issues have emerged or whether model retraining is required for adaptive improvement.**

**From a practitioner's perspective, this dual-purpose design facilitates continuous quality monitoring while maintaining throughput and interpretability demanded by regulated production. Limitations include dependence on consistent imaging conditions and potential sensitivity to lighting changes. Future work will explore adaptive attention mechanisms and federated retraining to maintain accuracy across production lines. Although demonstrated in medical-device manufacturing, the same lifecycle-monitoring principle can extend to other safety-critical inspection systems requiring explainable and auditable AI.**

***Index Terms*— Anomaly detection, automation systems, manufacturing systems, visual inspection systems, process monitoring, machine learning.**

## I. INTRODUCTION

AUTOMATED visual inspection is a cornerstone of modern manufacturing automation, particularly in highly regulated sectors such as medical devices, where every component must be verified for seal integrity, cleanliness, and sterility. Despite continuous advances in machine-vision technology, **human visual inspections (HVIs)** remain common and are estimated to account for nearly one-third of non-conformances in regulated production [1], [2]. The variability, fatigue, and subjectivity inherent to human inspection make it a major limitation within otherwise automated production lines, motivating a transition toward fully automated inspection systems that combine accuracy, consistency, and traceability. Inspection automation in medical-device manufacturing faces distinct challenges due to limited defect data, constrained hardware, and stringent process validation. Our previous work on sterile-barrier packaging quantified these limitations and demonstrated the need to re-engineer deep-learning inspection models for regulated production [2]. The Surface Seal Image (SSI) dataset [3] was developed to support this objective, and the MVTec-Zipper dataset [4] is used for cross-domain comparison.

Regulatory developments further emphasise the need for dependable and explainable inspection automation. As analysed in [5], the forthcoming **EU Artificial-Intelligence (AI) Act** classifies deep-learning inspection systems used in **Class III medical-device manufacturing** as *high-risk AI systems*. Such systems will be legally required to demonstrate full lifecycle traceability, documented risk management, and ongoing performance monitoring comparable to traditional manufacturing-software validation. These evolving obligations highlight the need for inspection frameworks that can support both real-time detection and long-term system validation.
Most academic anomaly-detection studies prioritise algorithmic generalisation across diverse datasets. While this goal advances generic AI research, it is misaligned with regulated automation, where conditions are fixed and defect types are known. To address this, our work focuses on deployability and process optimisation rather than unconstrained generalisation. The proposed dual-mode framework combines a **4-MS-SSIM** perceptual mode for inline detection with a **Mahalanobis-distance** mode for supervisory monitoring, together forming a single architecture that supports lifecycle traceability requirements [5]. The **dynamic attention-mask** strategy exploits known part geometry and fixture stability to restrict computation to relevant inspection areas, improving accuracy and speed under small-data conditions. Hardware limitations are incorporated as design parameters rather than constraints: all experiments use a single-GPU workstation typical of manufacturing cells. The **SSI dataset**, characterised by high resolution, high contrast, and low variability, reflects the imaging conditions of most real inspection stations and differs fundamentally from public datasets. These choices distinguish this work from more generic methods such as MOCCA [6], which target broad visual domains but require greater computational resources and less-controlled conditions.
**Contributions.** The core contributions of this work are as follows:

1. A **dual-mode, attention-guided autoencoder** that integrates 4-MS-SSIM inline anomaly detection with latent-space Mahalanobis monitoring, enabling both edge-level defect detection and supervisory lifecycle oversight within a single framework.
2. A **dynamic attention-mask mechanism** tailored to high-resolution, low-variability imagery that enhances efficiency and recall under small-data constraints.
3. A **deployment-oriented evaluation methodology** including explicit unsupervised and supervised threshold-selection strategies and experiments executed entirely under **single-GPU industrial hardware** representative of real production environments.

**Scientific Novelty in Automation Context.** Unlike previous studies that primarily advance visual-anomaly-detection algorithms, this work contributes to the automation science of quality assurance by defining a dual-mode inspection architecture that couples perceptual analysis with supervisory-level monitoring.

**Paper organisation**. Section II reviews related work; Section III describes the architecture and scoring modes; Section IV details datasets and implementation; Section V presents results; Section VI discusses automation implications; Section VII concludes the paper with final remarks and directions for future work.

## II. Related Work and Background

### *A. Deep Learning for Automated Inspection*

Over the past decade, deep learning (DL) has transformed computer vision by enabling automatic feature extraction from large image datasets. Convolutional Neural Networks (CNNs) have achieved state-of-the-art performance in classification and segmentation tasks across domains such as autonomous driving, biomedical imaging, and industrial defect detection. However, these successes generally rely on extensive and diverse class wise training data. In regulated manufacturing, especially for medical devices, such data abundance is rarely achievable because real defects occur infrequently, and product images exhibit low diversity. As demonstrated in our earlier study on sterile-barrier packaging [2], models trained on small, low-entropy datasets quickly overfit, limiting generalisation and undermining qualification for automated use in production.

To address these constraints, we have proposed compact CNN and autoencoder architectures to learn exclusively from **defect-free** samples – keeping in view to present an approach that is well aligned with automation requirements and explainable reconstruction-based inspection.

### *B. Anomaly Detection in Industrial Automation*

Anomaly detection—also referred to as novelty or outlier detection—aims to identify samples that deviate from expected normal behaviour. In manufacturing, where processes are highly repeatable and most products are conforming, anomaly detection provides a natural framework for automated inspection. Early methods relied on handcrafted features such as GFTT [21], MSER [22], FAST [23], BRIEF [24], and FREAK [25]. Later descriptors, including SIFT [26] and SURF [27], unified detection and description to improve rotation and scale invariance. With the emergence of deep learning, CNN-based descriptors such as DeepPatch [28], LIFT [29], and R2D2 [30] achieved far superior robustness and discriminative power compared with traditional approaches.

Among modern anomaly-detection paradigms, **Convolutional Autoencoders (CAEs)** are particularly more appealing because they learn compact non-linear representations of normal data [31]–[33]. CAEs provide an efficient means of dimensionality reduction and denoising, both valuable for industrial inspection. Variational Autoencoders (VAEs) [7], [34] extend CAEs with probabilistic latent variables, while Sparse and Stacked Autoencoders [32], [35], [36] increase representation capacity. Hybrid approaches further integrate reconstruction error with feature-space metrics or clustering objectives [36]. These techniques have been successfully applied to defect detection in domains such as textiles, additive manufacturing, and electronics assembly.

In **industrial automation**, imaging conditions are typically stable and repeatable—lighting, geometry, and part placement remain consistent across samples. This environmental stability allows unsupervised autoencoder-based methods to achieve robust performance even with small datasets. Recent review studies [35], [36] confirm that anomaly detection has become a mainstream strategy for automated quality control, enabling the identification of rare faults without explicit defect modelling. Consequently, our study builds on this research direction, focusing on lightweight, unsupervised architectures suitable for validation and deployment within highly regulated manufacturing systems such as medical-device production.

### *C. Deep Anomaly Detection Architectures*

Several state-of-the-art architectures were selected as baselines for comparison: the **Vanilla Autoencoder (AE)** [35]; **MOCCA**, a multilayer one-class classifier using centroid-based latent losses [6]; **CPCAE**, a patch-based CAE with positional encoding for improved localisation [37]; **RAG-PaDiM** [8], which extends PaDiM [38] with attention U-Nets [39]; and related AG-PaDiM variants. These models were re-implemented when public code was unavailable to ensure consistent baselines under identical data and hardware conditions.

These architectures represent the current **state of the art in anomaly detection**, and they were **re-implemented in this study** when public source code was unavailable. The objective was not to challenge their transferability to medical-device imagery but to establish **robust reference baselines** under the same industrial hardware and dataset constraints. Benchmarking these methods alongside our proposed dual-mode framework ensures a fair and transparent comparison of performance and resource efficiency across different anomaly-detection paradigms.

### *D. Small-Dataset Strategies for Manufacturing Automation*

Several strategies have been proposed to address data scarcity in manufacturing. **Transfer learning** from large datasets such as ImageNet can accelerate convergence [10]–[12], though domain differences limit its benefit for medical-device imagery. **Few-shot** and **meta-learning** models [13], [14] can adapt from small-labelled sets but are less effective when only defect-free data exist. Simple **Data-augmentation** methods [15], [16] may enlarge training sets but can degrade accuracy under tightly controlled imaging conditions as observed in previous work [2]. Recent research has also explored **generative data-synthesis** methods to address limited training data. **Generative Adversarial Networks (GANs)** such as *Defect-GAN* [17] can create and control synthetic defects, improving data diversity. However, GANs require many defective examples for stable training, which is impractical in medical-device manufacturing where true anomalies are scarce. More recently, stable **diffusion models** have emerged as an alternative that requires less data while providing high-fidelity synthesis. For example, *SynAdult* [18] demonstrated multimodal dataset generation via diffusion models and neuromorphic simulation for biometric applications. Although diffusion models show promise, control over defect morphology remains limited and may need random generation followed by post-classification to select usable synthetic samples. These techniques represent a promising research direction for future work but remain beyond the scope of this study.

### *E. Automation and Regulatory Perspective*

Although deep anomaly-detection research is extensive, few studies address automation-system validation and lifecycle assurance. Industrial automation requires not only accurate but also verifiable and maintainable AI systems. The forthcoming **EU Artificial Intelligence Act** [5] introduces new obligations for continuous monitoring and traceable performance, complementing established quality-system standards such as **ISO 13485** and **ISO/IEC 42001**. Recent T-ASE studies on intelligent visual inspection and lifecycle monitoring [19], [20] focus mainly on algorithmic or predictive-control integration in general manufacturing contexts. In contrast, this work introduces a **dual-mode inspection framework** that combines perceptual anomaly detection with statistical lifecycle monitoring, aligning automation-system design with traceability and validation requirements for regulated industrial environments.

## III. Proposed Methods and Framework

### *A. Overview of the Dual-Mode Architecture*

The proposed framework builds on deep anomaly-detection principles but adapts them to the constraints of regulated **medical-device manufacturing**. It integrates two complementary scoring modes within a single deployable architecture: (1) a **structural-similarity-based mode** for real-time inline detection and (2) a **feature-distance-based mode** for supervisory lifecycle monitoring.

Both modes share an **attention-guided autoencoder backbone** trained on defect-free samples, enabling a single model to perform edge inference and supervisory monitoring without retraining.

### *B* Dynamic *Attention-Mask Strategy and Autoencoder Backbone*

Inspection images are typically high-resolution and geometrically consistent, which allows the use of a lightweight **attention-mask** to extract only defect-prone regions of interest (ROIs). The approach follows established machine-vision practice in regulated environments, where inspection zones are predefined by risk assessment, as in the SSI and MVTec-Zipper datasets.

While the specific function design of the attention mask may vary based on the domain-specific requirements of the datasets, for the scope of these experiments we implement a fundamental function computed in three stages: **Edge-preserving smoothing** using bilateral filtering [41]; **Gradient-based edge** **detection** (Canny operator [42]) to locate material boundaries; and **Histogram-profile analysis** along vertical or horizontal axes to refine ROI coordinates.

Algorithm 1: Attention-Mask extractor function for the SSI and MVTec-Zipper datasets

**Input**: dataset of images $\mathcal{D} = \{I_i\}_{i=1}^{N}$, edge search direction $\mathbf{d} \in \{V\uparrow, V\downarrow, H\rightarrow, H\leftarrow\}$, patch size $\boldsymbol{p}$ = ($p_h$, $p_w$)
**Output**: Attention-Mask Coordinates $R_i$={($x_0$, $y_0$), ($x_1$, $y_1$)}

1 **for** all **i** in **N**
2 # Step 1 - Edge Preserving Smoothing
3 $I_{smooth}$ ← `Bilateral Filter(Ij, σs, σr)`
`▷ Suppress noise while preserving material boundaries`
4 #Step 2 – Gradient-Based Edge Detection
5 E ← `CannyEdgeDetector(Ismooth, Tlow, Thigh)`
`▷ Locate candidate edges corresponding to seal or fabric boundaries`
6 #Step 3 – Histogram-Profile Refinement
7 **If** d **in** $\{V\uparrow, V\downarrow\}$ **then**
8 $H_x$ ← `HorizontalProjection(E)`
9 ($x_0$, $x_1$) ← ExtractPeaks ($H_x$)
10 **Elseif** d **in** $\{H\rightarrow, H\leftarrow\}$
11 $H_y$ ← `VerticalProjection(E)`
12 ($y_0$, $y_1$) ← ExtractPeaks ($H_y$)
`▷ Identify boundary coordinates enclosing defect-prone zones`
13 #Step4 – Patch Definition
14 $R_i$ ← `CropRegion(Ii,` $\boldsymbol{p}$, {($x_0$, $y_0$), ($x_1$, $y_1$)})
15 `Return Set {Ri} for all i`

For the SSI dataset, a single Patch per image is sufficient since all defects occur along the seal line, whereas the MVTec-Zipper dataset requires three Patches corresponding to the central zip and the fabric edges.

The autoencoder backbone is a **four-layer convolutional encoder–decoder** network trained on defect-free samples only, using **mean-squared-error (MSE)** as the reconstruction loss. Training is unsupervised, reflecting realistic production conditions where defects are rare. The network output is a reconstructed attention patch $\hat{I}_j$ approximating the input $I_j$.

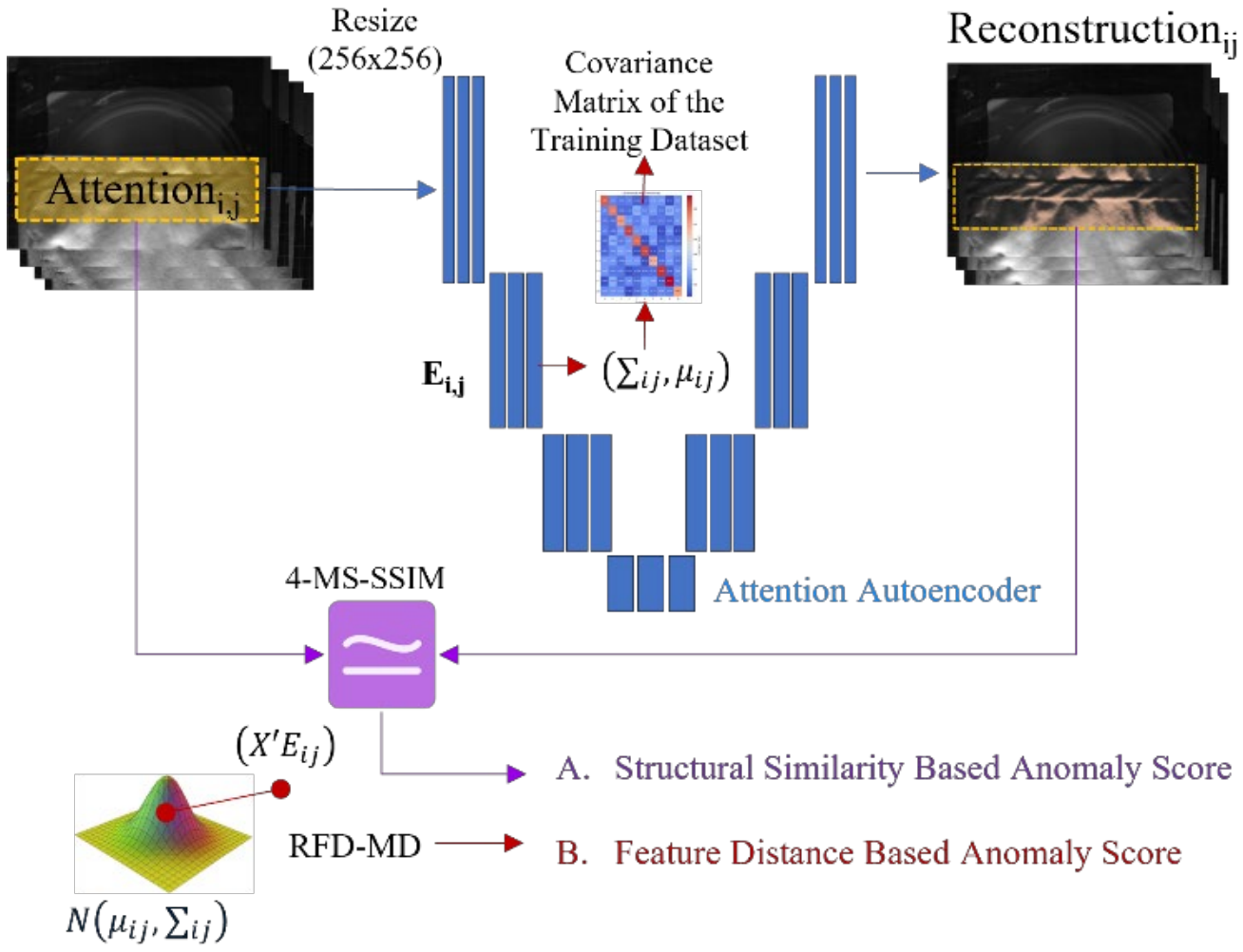

**Fig. 1.** Illustration of the proposed overall architecture. The shared autoencoder feeds two independent scoring modules: the structural-similarity-based module and the feature-distance-based module.

### *C. Structural-Similarity-Based Anomaly Scoring*

Once the autoencoder has been trained on normal data, inference proceeds by comparing each input patch with its reconstruction using the **structural-similarity index (SSIM)** [43]. In contrast to prior anomaly detection approaches that integrate SSIM into the training loss, our framework employs it solely during inference to evaluate reconstruction quality. Although this design choice may reduce the autoencoder's generalisation ability when reconstructing unseen samples, it effectively preserves its discriminative power between normal and anomalous patterns. This is because incorporating SSIM in the loss function often promotes over-generalisation, enabling the model to reconstruct defective regions too accurately and thereby diminishing defect separability.

To further increase perceptual sensitivity, we employ the **4-MS-SSIM** variant introduced by Renieblas et al. [44], which aggregates four similarity components—luminance, contrast, and structure—across multiple scales. This specific scoring index was selected based on a study comparing SSIM variants as well as other more common scoring indexes such as MSE and MAE. The detailed results of this study are summarized in the project repository. Although 4-MS-SSIM has demonstrated superior discrimination in radiological imaging, to the best of our knowledge it has not previously been applied to anomaly detection. Its multi-scale, gradient-enhanced formulation is particularly effective for high-resolution industrial imagery, where subtle structural distortions along seal boundaries signify true defects.

Anomaly scores are computed as $L_i = 1 - SSIM_{4-MS}(\hat{I}_j, I_j)$. Thresholds are defined either unsupervised $\boldsymbol{T_u}$, for use in early deployment when limited deffect data is available, or supervised $\boldsymbol{T_s}$, for use in post-qualification stages with abundant deffect data.

$$T_U = \mu_{norm} + 2\sigma_{norm} \quad (1)$$

where $\mu_{norm}$ and $\sigma_{norm}$ represent the mean and standard deviation of the normal scores, covering approximately 97.7% of the normal-score distribution, providing a statistically interpretable decision boundary for inspection systems.

$$T_S = \underset{T\in[0,1]}{\operatorname{argmax}} \frac{1}{N}\sum_{i=1}^{N} 1[(\mathcal{L}_i \geq T) \Leftrightarrow (y_i = 1)] \quad (2)\ (1)$$

where $y_i \in \{0,1\}$ indicates the ground-truth label.

### D. Feature-Distance-Based Anomaly Scoring

The second mode detects anomalies in the **latent-feature space**, similar to MOCCA [6] and RAG-PaDiM [8], by modelling feature embeddings as multivariate Gaussians.

Let $E_{ij}$ represent the embedding of patch $i$ from layer $j$. The **Mahalanobis** distance between $E_{ij}$ and the reference distribution $N\left(\mu_{ij}, \Sigma_{ij}\right)$ is

$$D_M(E_{ij}) = \sqrt{(E_{ij} - \mu_{ij})^T \Sigma_{ij}^{-1} (E_{ij} - \mu_{ij})} \quad (3)$$

The resulting Mahalanobis distances are larger for anomalies. To reduce memory and computation, a **random feature-drop** strategy samples 500 latent dimensions while maintaining discriminative power (Johnson–Lindenstrauss principle).

Thresholds follow the same supervised and unsupervised principles as Eq. (1)– (2). In industrial use, the **Random Feature Drop with Mahalanobis Distance** (RFD-MD) mode supports statistical drift monitoring and may be used to trigger retraining as additional data accumulate.

### E. Thresholding Rationale for Regulated Automation

Threshold definition is a critical element of system qualification in regulated inspection environments. Many anomaly-detection studies report only global metrics such as AUC or maximum theoretical accuracy, which rely on post-hoc optimisation and are unsuitable for validated production software. In contrast, the proposed approach defines thresholds *a priori* to ensure repeatability and traceability. The unsupervised rule $\mu_{norm} + 2\sigma_{norm}$ represents a well-established 95–99% confidence interval that can be justified under ISO 13485 process-validation guidelines. The supervised variant reflects the re-qualification step performed when new defect data become available during production. Together, these deterministic thresholds align model behaviour with regulatory expectations for traceability, documentation, and lifecycle validation, contributing to the engineering-validation framework discussed in Section V-F.

### F. Computational and Hardware Considerations

All experiments were performed on a single RTX 4080 GPU with an Intel i7 CPU and 64 GB RAM. Software packages and versioning information is detailed in the project repository.

## IV. Experimental Design

The objective of this experimental study is to **validate the proposed dual-mode architecture under realistic industrial and regulatory conditions**, evaluating its robustness and deployability relative to representative state-of-the-art (SotA) anomaly-detection methods.

### A. Datasets

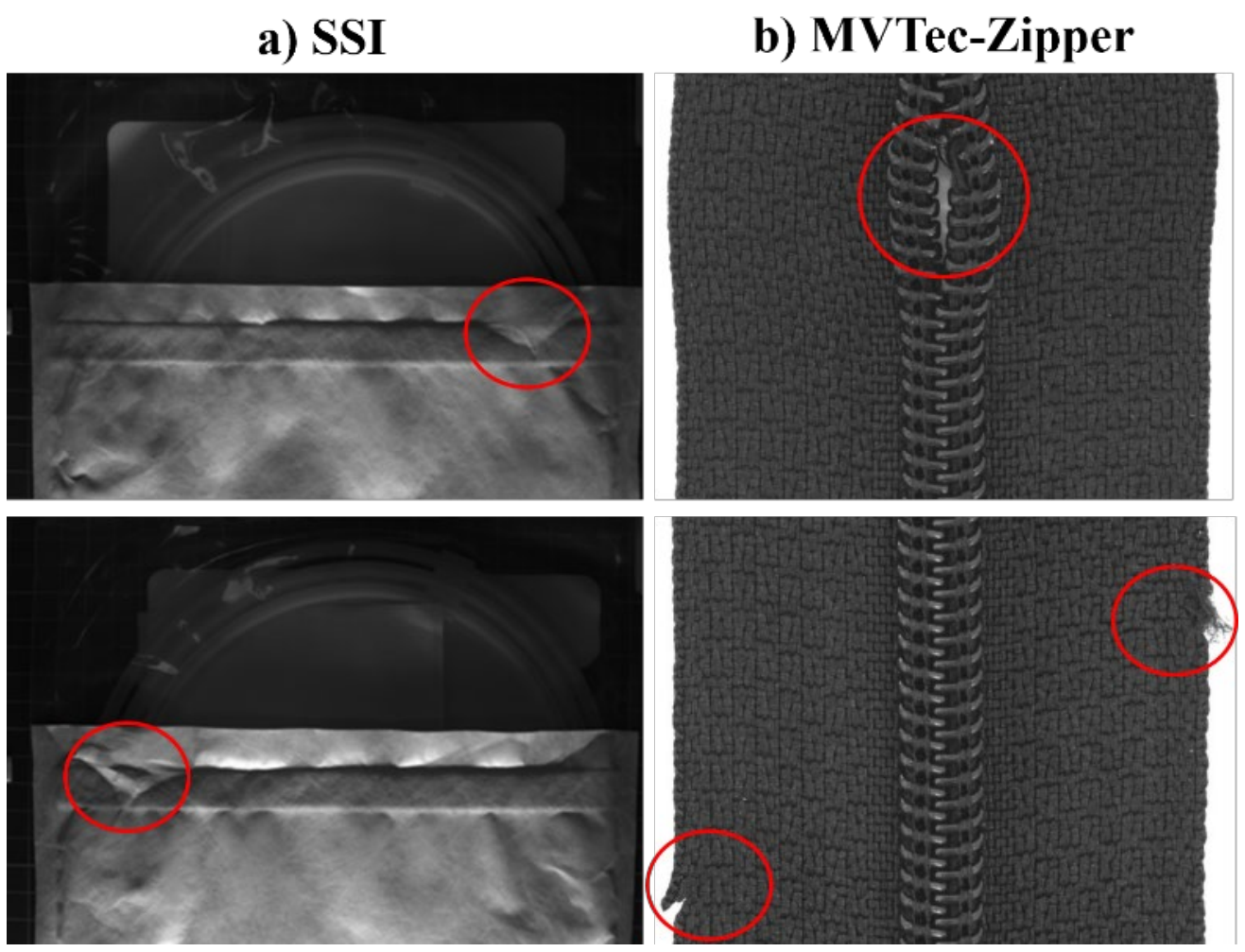

**Fig. 2.** dataset example images. a)-left images corresponding to the SSI dataset, b)-right images corresponding to the MVTec-Zipper dataset. Top image shows a defect over the zipper; bottom image shows a defect in the fabric.

1) **Surface Seal Image (SSI) Dataset**
Experiments used the **Surface Seal Image (SSI)** dataset [3], comprising ~1,200 grayscale (3008x4110 pixels) images captured under factory conditions. It includes both normal and defective samples and represents the controlled, low-variability imagery typical of medical-device inspection.

2) **MVTec-Zipper Dataset**
Cross-domain validation employed the **MVTec AD Zipper** [4], consisting of 391monochorm (1024x1024 pixels) images. Although outside the medical domain, this dataset offers a controlled, high-resolution benchmark for structural defects. It enables direct comparison with published results such as **MOCCA** and supports evaluation of model generalisation across domains. Balanced Accuracy (BA) served as the primary metric to compensate for class imbalance and aligned with published results.

### B. Baseline Models and Benchmarking Protocol

For comparison, we implemented representative state-of-the-art models under identical image resolution and hardware settings: the **Vanilla Autoencoder** [35], **MOCCA** [6], **CPCAE** [37], and **PaDiM**-based variants including **RAG-PaDiM** [8] and **AG-PaDiM** [38], [39]. When public code was unavailable, implementations followed published specifications to ensure consistent baselines. These are reported as *reference re-implementations* rather than strict reproductions, and all source code has been released for verification. Each experiment was repeated three times with fixed random seeds to mitigate stochastic effects.

### C. Implementation Details and Optimisation

Code and interim ablation results are available in the public repository: https://github.com/JulioZanon/Dual-Mode-Deep-Anomaly-Detection-for-Medical-Manufacturing.
Model optimisation followed a consistent two-stage workflow using Bayesian optimisers, as detailed in the repository. Training employed an 80/20 split between training and cross-validation subsets with fixed random seeds, and the mean and standard deviation across three runs were reported for all evaluation metrics.

### D. Evaluation Metrics

Evaluation metrics reflected both algorithmic performance and operational reliability: **Accuracy (ACC)**, **Precision (P)**, **Recall (R)**, and **F1-score (F1)** characterised classification behaviour; **AUC** measured discrimination independent of a fixed threshold; **Balanced Accuracy (BA)** supported comparison with published MVTec results; and **False-Negative Rate (FNR)** quantified the risk of undetected defects. For the SSI experiments, both supervised and unsupervised thresholds (Section III-E) were assessed to capture early-deployment and post-qualification behaviour, a key aspect of lifecycle validation under the EU AI Act [5].

### E. Experimental Objectives

The overall experiments were designed to address three research questions relevant to manufacturing automation:

1. **Baseline comparison:** How does the proposed dual-mode architecture perform relative to established SotA methods under the same data and hardware constraints?
2. **Threshold behaviour:** How do unsupervised and supervised thresholding regimes influence recall and false-reject rates when defective samples are scarce?
3. **Generalisation vs. deployability:** What trade-offs arise between cross-domain accuracy (MVTec) and domain-specific reliability (SSI)?

By explicitly linking these questions to industrial and regulatory priorities, the experimental framework supports a balanced evaluation of algorithmic performance, interpretability, and deployability in automation contexts.

## V. Engineering Validation and Results

The experimental results are divided and presented into four parts.

- First, we report baseline performance on the **SSI dataset** to establish the relative accuracy of the proposed methods against SotA architectures. Second, we evaluate thresholding behaviour under different levels of defect availability.
- Third, we examine the results of the **feature-distance-based** mode and its suitability for supervisory monitoring.
- Finally, we analyse **cross-domain generalisation** using the **MVTec-Zipper** dataset and discuss the implications of these findings for automation and regulatory compliance.

### A. Industrial Validation Study on the SSI Dataset

Table 1 summarizes the results of the industrial validation on the SSI test set. Under identical resolution (256 × 256 pixels) and hardware conditions, the proposed **4-MS-SSIM** method achieved the highest overall performance, followed by the **RFD-MD** configuration. Both methods outperformed established baselines, including **CPCAE** and **MOCCA**, demonstrating superior accuracy and robustness while maintaining low computational cost. The **Vanilla Autoencoder**, which forms the basis of our approach, confirmed that reconstruction-only metrics are inadequate for small, low-variability datasets.

TABLE 1. BENCHMARK STUDY TRAINED WITH SSI AND RESULTS OBTAINED FROM THE SSI TEST SUBSET (ORDER BY DESCENDING ACC)

| Model | AUC | ACC | P | R | F1 |
|---|---|---|---|---|---|
| [b]4-MS-SSIM | 0.977 | 0.931 | 0.938 | 0.922 | 0.930 |
| [b]RFD-MD | 0.927 | 0.875 | 0.833 | 0.939 | 0.883 |
| [b]CPCAE | 0.817 | 0.742 | 0.664 | 0.978 | 0.791 |
| [b]MOCCA GPU Adaptation | 0.724 | 0.694 | 0.639 | 0.894 | 0.745 |
| [b]Padim | 0.720 | 0.664 | 0.620 | 0.844 | 0.716 |
| [a]MOCCA Original | 0.676 | 0.634 | 0.679 | 0.528 | 0.594 |
| [b]Vanilla AE (4 Layers) | 0.614 | 0.614 | 0.936 | 0.244 | 0.388 |
| [b]AG-Padim | 0.664 | 0.598 | 1.000 | 0.748 | 0.664 |
| [b]RAG-Padim | 0.500 | 0.500 | 1.000 | 0.667 | 0.500 |

**a:** *Original code trained with downsized images to overcome memory limitations (1024x1024)*
**b:** *GPU adaptation trained with full resolution (3008x4110).*

### *B. Thresholding Behaviour and Defect Availability*

Table 2 compares supervised and unsupervised thresholding. For structural-similarity mode, accuracy under unsupervised calibration (ACC 0.903) was close to the supervised case (ACC 0.931), confirming that statistical thresholds provide reliable performance even without labelled defects. The feature-distance mode showed greater sensitivity to threshold choice, yet both configurations maintained high recall. These results support a lifecycle strategy in which an inspection system initially operates with unsupervised limits and later transitions to supervised re-qualification as defect data become available.

TABLE 2. SUPERVISED VS UNSUPERVISED THRESHOLD STRATEGY COMPARISON

| Model | ACC | P | R | F1 |
|---|---|---|---|---|
| 4-MS-SSIM – Supervised | 0.931 | 0.938 | 0.922 | 0.930 |
| 4-MS-SSIM - Unsupervised | 0.903 | 0.847 | 0.983 | 0.910 |
| RFD-MD – Supervised | 0.722 | 0.643 | 1.000 | 0.783 |
| RFD-MD - Unsupervised | 0.658 | 0.594 | 1.000 | 0.745 |

### *C. Feature-Distance-Based Scoring and Latent-Space Monitoring*

Several dimensionality-reduction methods were explored, including PCA, ICA, and random feature selection; full results of ablation studies are provided in the repository. Among these, random feature-drop combined with Mahalanobis distance achieved the best balance of accuracy and efficiency (ACC 0.836, F1 0.848). Despite slightly lower precision than the inline 4-MS-SSIM detector, this latent-space approach proved valuable for continuous monitoring. Its sensitivity to small variations in normal samples enables detection of process drift, sensor degradation, or illumination changes, supporting the post-deployment validation requirements of the **EU AI Act** [5].

### *D. Cross-Domain Evaluation on MVTec-Zipper*

Table 3 reports the cross-domain results on the **MVTec-Zipper** dataset. The structural-similarity mode achieved a **MaxBA of 0.742**, close to the published **MOCCA** value (0.78) and above other baseline methods (AESSIM 0.70, AEL2 0.689). Although we could not reproduce MOCCA's reported accuracy using its public repository—which lacks pretrained weights and detailed training guidance—our reproduced results were consistently lower, suggesting substantial dataset-specific tuning may be required. In contrast, the proposed model generalised from the SSI configuration with only minor retraining, demonstrating strong robustness without dataset-specific optimisation.

The feature-distance method attained lower accuracy (ACC 0.55), consistent with its sensitivity to domain-specific feature statistics and restricted by a very low number of normal samples to train with only 240 images.

TABLE 3. MVTEC-ZIPPER DATASET RESULTS COMPARING OUR PROPOSED APPROACH WITH PUBLIC RESULTS

| Model | MaxBA |
|---|---|
| Our method with anomaly score based on **4-MS-SSIM** (all images) | 0.742 |
| Our method with anomaly score based on **Mahalanobis distances** (all Images) | 0.55 |
| [a]MOCCA | 0.78 |
| [b]MVTec $AE_{SSIM}$ | 0.7 |
| [b]MVTec $AE_{L2}$ | 0.689 |
| [b]MVTec AutoGan | 0.566 |
| [b]MVTec CNN Feature Dictionary | 0.531 |

**a**: *Results reported in* [6]; **b**: *Results reported in* [4].

### *E. Comparative Discussion and Automation Relevance*

Across all experiments, the proposed dual-mode architecture consistently outperformed or matched SotA methods while operating within realistic industrial constraints.

Three aspects underpin this result:

1. **Attention-guided efficiency:** ROI masking limits computation to defect-prone areas, improving speed and recall under small-data conditions.
2. **Perceptual sensitivity:** The 4-MS-SSIM metric enhances anomaly separability without retraining, offering interpretable scoring that can be documented within quality-system records.
3. **Lifecycle monitoring:** Latent-space Mahalanobis scoring enables ongoing statistical surveillance of process drift, supporting continuous validation and traceability under regulatory frameworks such as ISO 13485 and the EU AI Act [5].

From an automation perspective, these attributes translate directly into operational value.

The inline SSIM-based mode provides real-time defect detection at line rate, while the feature-distance mode enables supervisory oversight and performance auditing across production batches.

Together, they form a complete inspection-automation framework that satisfies both throughput and compliance requirements—key pillars of the **Automation Science and Engineering** domain.

### *F. System Validation and Reproducibility*

The proposed framework was evaluated following an engineering-validation approach consistent with good manufacturing-practice expectations for software tools affecting product quality. Key aspects of this validation include:

• **Deterministic thresholding** (Section III-E): fixed, statistically interpretable decision limits ($\mu + 2\sigma$ or supervised calibration) defined prior to testing to ensure traceability.

• **Controlled hardware configuration** (Section III-F): all experiments executed on identical single-GPU systems representative of current industrial vision-cell hardware, with inference times of approximately 50 ms per ROI.
• **Repeatability across runs:** each test repeated three times with fixed random seeds; results reported as mean ± standard deviation.
• **Alignment with ISO 13485 validation principles:** explicit documentation of configuration, reproducible datasets, and locked thresholds mirrors software-tool validation practices in regulated manufacturing, including the open-source code repository.
Collectively, these elements establish the repeatability and traceability required for engineering validation of automated inspection systems.

These validation outcomes form the foundation for the broader automation implications discussed in Section VI.

## VI. Automation Science And Engineering Implications

The results presented in Section V demonstrate that the proposed dual-mode deep-anomaly-detection framework deployed as depicted in Figure 3, delivers competitive accuracy while satisfying the operational and traceability needs of industrial automation. Beyond quantitative performance, several broader implications emerge concerning **system integration, lifecycle governance, and regulatory readiness** for AI-enabled inspection systems in medical-device manufacturing.

This work also advances the **science and engineering of automation** by demonstrating how deep-learning subsystems can embody key automation-theory principles. The dual-mode framework improves **reliability** through redundant scoring channels, enhances **observability** by translating latent-space features into interpretable metrics, and supports **control of process drift** via continuous feedback of feature-distribution statistics to supervisory systems. These capabilities enable integration with higher-level **supervisory control and data-acquisition (SCADA)** or manufacturing-execution systems, thereby connecting perception-level intelligence with closed-loop automation control.

### *A. From Laboratory Prototypes to Production-Line Automation*

Deep-learning-based inspection systems often remain confined to research environments because of limited interpretability, large computational requirements, or the absence of qualification pathways within established manufacturing standards. The proposed architecture addresses these barriers through its **attention-guided design** and **explicit thresholding procedures**, which facilitate deployment on standard industrial hardware. Both anomaly-scoring modes are computationally lightweight and compatible with common vision-cell controllers, enabling retrofit into existing inspection stations without major hardware upgrades.

By supporting real-time inference (≈ 50 ms per ROI) and providing quantitative decision boundaries, the system can be validated in accordance with **ISO 13485** and **FDA QSR** requirements for software tools affecting product quality. This bridges the gap between research prototypes and production-ready automation modules.

### *B. Lifecycle Monitoring and Regulatory Alignment*

As discussed in [5], the **EU AI Act** introduces new obligations for high-risk AI systems used in Class III medical-device manufacturing, including requirements for post-deployment monitoring, performance-drift detection, and documented risk management. The proposed dual-mode framework provides a direct technical pathway for **engineering validation** of these capabilities, supporting both lifecycle monitoring and regulatory alignment under standards such as **ISO 13485** and the forthcoming **ISO/IEC 42001**.

The **structural-similarity mode** ensures reliable inline operation and supports explainability through reconstruction-error overlays, while the **feature-distance mode** enables statistical monitoring of latent-space stability over time, allowing early detection of data drift or environmental variation. Together, these mechanisms realize the concept of **continuous validation**, linking automated inspection with lifecycle-management practices consistent with emerging AI-system governance frameworks.

### *C. Explainability and Human-in-the-Loop Oversight*

Explainability remains a cornerstone of automation adoption in regulated domains. Unlike black-box classifiers, the SSIM-based approach offers a clear physical interpretation of its decisions: anomalies correspond to local structural discrepancies visible in the reconstructed image. This facilitates **human-in-the-loop verification**, where quality engineers can visualize, audit, and approve inspection outcomes using interpretable metrics. Such transparency reduces the cognitive gap between automated and manual inspections, supporting operator trust and regulatory acceptance.

### *D. Integration into Automation Workflows*

In practice, modern manufacturing systems combine multiple automation layers—from inline defect detection to cloud-based analytics as depicted in Figure 3. The dual-mode framework naturally maps onto this hierarchy:

- The **SSIM-based mode** operates at the *edge* for real-time defect rejection.
- The **Mahalanobis-based mode** functions at the *supervisory* or *cloud* level, aggregating feature statistics across batches for process optimization and long-term performance tracking.

Each inspection computer within the production environment operates an application developed in **.NET** *(+MQTTnet)* or **Python** *(+Eclipse Paho)* that transmits latent feature data at fixed intervals—one per part, with typically cycle times in medical device manufacturing of one minute. Each message contains five hundred *float32* features per patch that describe the inspected component. The data are transmitted over the **MQTT** protocol using **TLS** encryption with mutual certificate authentication to ensure confidentiality and integrity. Messages are serialised in **Protocol Buffers**, producing compact payloads of roughly 2 KB, and are published with reliable delivery to a **cloud ingestion broker**, such as *AWS IoT Core* or an equivalent managed MQTT service. The broker authenticates devices and routes incoming data to a **stream-processing layer**—for example, a managed streaming platform such as *Kinesis Data Streams* or *Apache Kafka*. Containerised **analytics services** hosted in a cloud orchestration environment, such as *ECS* or *Kubernetes*, consume the stream, validate incoming records, and compute the **Mahalanobis distance** between each feature vector and a multivariate Gaussian model representing normal operational conditions. Computed analytics and metadata are stored in a **columnar OLAP database** optimised for analytical workloads and time-series queries—such as *Amazon Redshift* or *ClickHouse*—to support efficient large-scale aggregation, feature drift detection, and historical trend analysis. Detected anomalies or distributional shifts trigger alert events, which are propagated through a **cloud-based event and notification service** (for instance, *EventBridge* and *SNS*) to notify operators via e-mail, dashboards, or system integrations.

This architecture allows continuous feedback between production and quality-assurance systems, enabling **adaptive maintenance** and **data-driven process improvement** without violating static-model validation requirements.

### *E. Industrial Deployment Considerations*

Deploying AI-based inspection within validated manufacturing environments requires careful attention to reproducibility, cybersecurity, and data governance. The proposed system mitigates these concerns by:

1. Operating on locally stored ROI data to minimize network dependency; while upstreaming latent features vectors with significantly reduced payload.
2. Using deterministic threshold settings that can be locked and version-controlled; and
3. Providing auditable performance logs suitable for inclusion in device-history records.

It is worth noting that large-scale adoption of AI-driven visual inspection is constrained by storage requirements. For example, a manufacturing facility with 500 stations capturing one 20 MP image per minute over two shifts would produce roughly 275 TB of data per month. In contrast, storing only latent feature vectors of 2 KB per image would reduce the monthly storage demand to about 27 GB.

### *F. Broader Impact on Automation Science*

From an automation-science perspective, this work exemplifies a paradigm shift from pure algorithmic novelty to **systems-level innovation**—combining perception, decision making, and governance within a deployable inspection framework. The proposed dual-mode design demonstrates how deep learning can be engineered for *reliability* and *traceability*, transforming anomaly detection from a laboratory exercise into a validated automation component. As similar regulatory frameworks emerge globally, the principles demonstrated here—explicit thresholding, lifecycle monitoring, and explainable scoring—are expected to inform the design of next-generation intelligent automation systems across diverse manufacturing sectors.

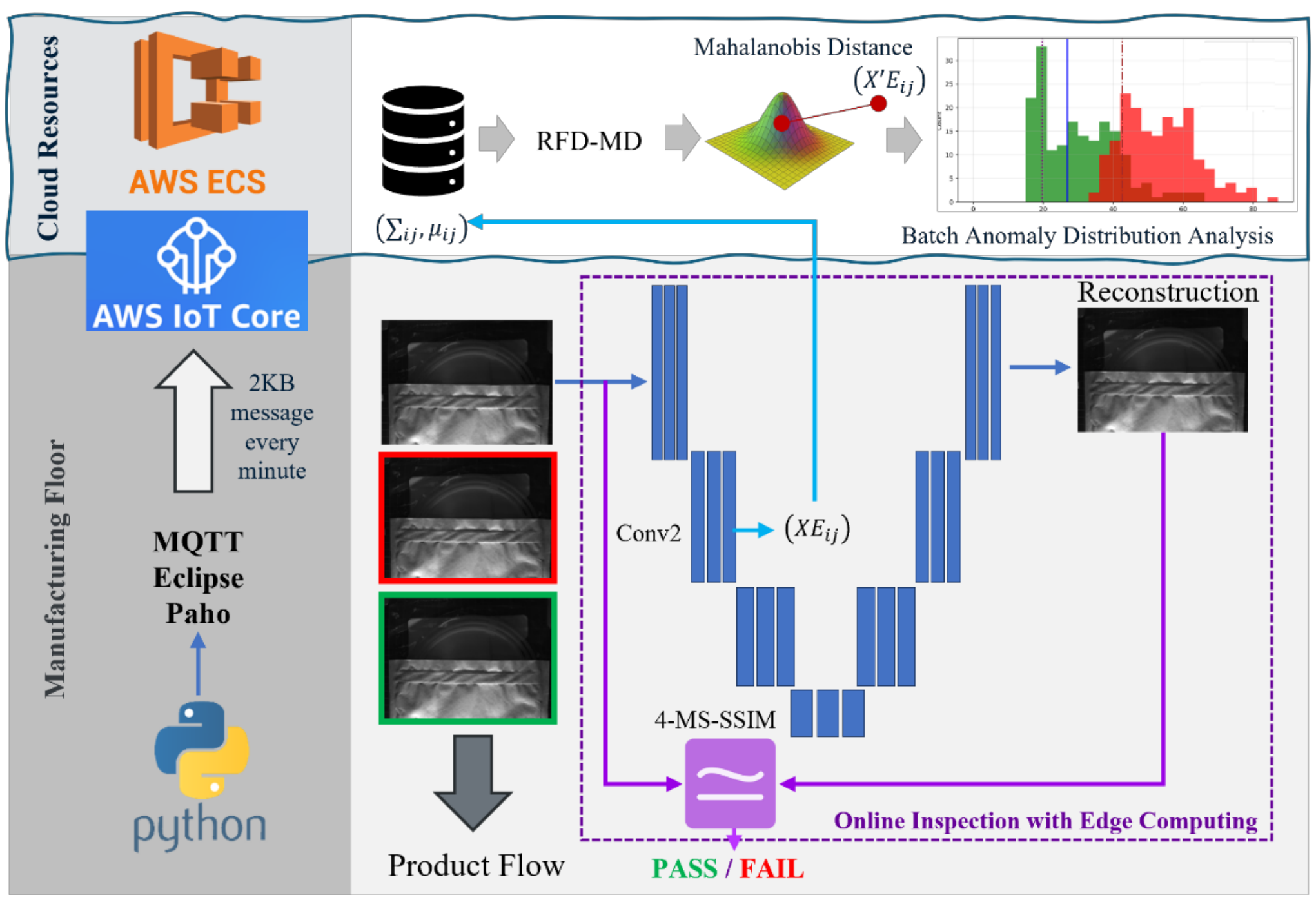

**Fig. 3.** Practical deployment framework for the proposed anomaly detection methods clarifying the automation hierarchy (edge vs supervisory) and based on AWS environment as an example.

## VII. Conclusion and Future Work

This paper presented a **dual-mode deep-anomaly-detection framework** designed for the automation of visual inspections in medical-device manufacturing. The approach integrates two complementary strategies—a **structural-similarity-based mode** for inline inspection and a **feature-distance-based mode** for supervisory monitoring—within a single attention-guided autoencoder backbone. By combining interpretable perceptual scoring with latent-space monitoring, the framework addresses both operational efficiency and lifecycle traceability, two critical requirements for deploying AI in regulated automation.

Experimental results on the **Surface Seal Image (SSI)** dataset confirmed that the proposed structural-similarity method (**4-MS-SSIM**) outperforms recent state-of-the-art architectures, achieving high accuracy on limited data and constrained hardware. The Mahalanobis-based feature-distance mode (**RFD-MF**) demonstrated its value as a lightweight monitoring layer for tracking process drift and environmental variation. Together, the two modes form a deployable inspection architecture capable of satisfying both real-time production demands and emerging compliance expectations under the **EU AI Act** and **FDA QSR** frameworks.

From a broader automation-science standpoint, this work underscores the importance of **system-level design** that integrates algorithmic performance with explainability, threshold governance, and reproducibility. The results show that careful engineering choices—such as decoupling SSIM from the loss function, adopting an ROI-driven attention mechanism, and constraining models to single-GPU execution—can yield AI systems that are not only accurate but also verifiable and auditable in production environments.

**Future work** will extend this foundation in several directions:

1. **Full-image attention overlays** and adaptive loss weighting to detect defects beyond predefined ROIs while optimising computational requirements;
2. **Hybrid training objectives** that combine MSE and SSIM terms to balance reconstruction fidelity and perceptual sensitivity;
3. **Advanced feature-reduction techniques** such as attention-based ranking or random-projection ensembles to enhance efficiency while preserving discriminative power;
4. **Validation on larger, predominantly normal datasets** to exploit intentional over-fitting to the defect-free class and improve generalization within controlled domains;
5. **Expansion of domain-specific repositories** to address the scarcity of public datasets for medical-device inspection;
6. **Development of regulatory alignment protocols**, including model-release dossiers and qualification templates consistent with the EU AI Act; and
7. **Replication of resource-intensive architectures** such as MOCCA on next-generation hardware to quantify the trade-offs between memory demand and accuracy at native image resolution.

Ultimately, the proposed architecture demonstrates that deep learning can be engineered for **automation reliability, interpretability, and lifecycle governance**, advancing the integration of AI into safety-critical manufacturing processes and contributing to the foundational methods of **Automation Science and Engineering**.

**Automation Impact.** By unifying perceptual anomaly detection and statistical lifecycle monitoring within a single architecture, this work **illustrates a transferable automation framework** for integrating explainable AI into closed-loop quality control. The results highlight how AI-driven perception and supervisory analytics can coexist within validated production environments, suggesting a new design pattern for next-generation automation systems. In this context, the study advances the field of **Automation Science and Engineering** by **moving regulatory traceability toward an intrinsic system attribute rather than an external compliance layer**, providing a practical direction for future research on regulation-aware automation architectures.

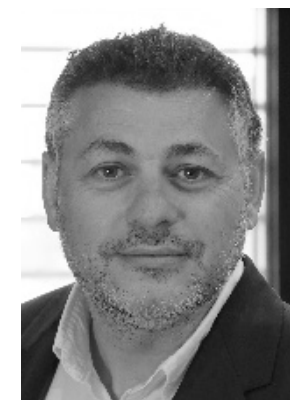

**Julio Zanon Diaz** received his degree in Telecommunications Engineering from the University of Valencia, Spain, in 2000, prior to the Bologna Process reforms that aligned third-level degrees across Europe. He later returned to the University of Valencia to complete a B.Eng. (Hons.) in Electronics in 2017. Since 2001, he has worked as an automation engineer in the automotive and medical device industries, gaining recognised expertise in equipment design, safety, software development, robotics, and machine vision. He is currently a Fellow Engineer at Boston Scientific, where he leads strategies and teams in the development and deployment of advanced manufacturing technologies, with a focus on data science, artificial intelligence, digital twins, and human-centric manufacturing. Mr. Zanon Diaz is also pursuing a Ph.D. in Electrical and Electronic Engineering at the University of Galway, Ireland. His research interests include automated visual inspection using computer vision and deep neural networks in highly regulated manufacturing environments. He previously served as a part-time Adjunct Lecturer with the University of Galway.

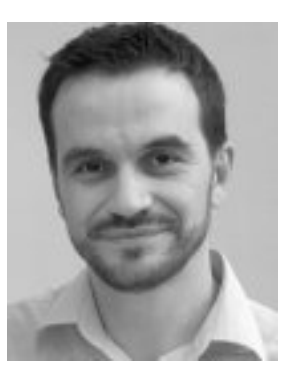

**George Siogkas (Member, IEEE)** is a Senior R&D Consultant and the founder of Irish company CVRLab, specializing in Computer Vision and Machine Learning. He has worked as a R&D Computer Vision engineer for leading companies in the fields of Autonomous Driving and Medical Devices and is currently working with Boston Scientific. He holds a PhD in Electrical and Computer Engineering from the University of Patras, Greece, as well as an MSc and MEng in related fields. Throughout his career, he has combined academic roles with industry positions focused on applied research and development in Computer Vision and Pattern Recognition and has contributed to advancing methods in Computer Vision both through research published in peer-reviewed international journals and conferences and through cutting-edge research for product development, including projects at various stages from concept to commercialization.

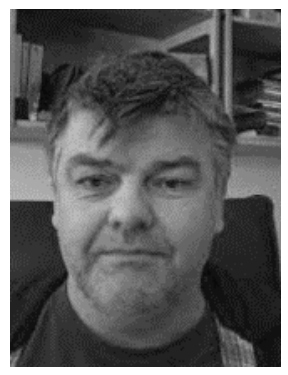

**Peter Corcoran** (Fellow, IEEE) currently holds the Personal Chair of Electronic Engineering with the College of Science and Engineering, University of Galway. He was the Co-Founder of several start-up companies, notably FotoNation (currently the Imaging Division, Xperi Corporation). He has more than 600 cited technical publications and patents, more than 120 peer-reviewed journal articles, 160 international conference papers, and a co-inventor on more than 300 granted U.S. patents. He is an IEEE fellow recognized for his contributions to digital camera technologies, notably in-camera red-eye correction and facial detection. He is also a member of the IEEE Consumer Technology Society for more than 25 years and the Founding Editor of IEEE Consumer Electronics Magazine.